\documentclass[a4paper,11pt]{article}

\usepackage[mode=build]{standalone}

\usepackage{fourier}
\usepackage{color}
 \usepackage{graphicx}
\usepackage{url}
\usepackage[affil-it]{authblk}
\usepackage{amsmath}
\usepackage{wrapfig}

\usepackage[T1]{fontenc}
\usepackage{times}

\usepackage{natbib}
\usepackage{apalike}

\usepackage{hyperref}
\usepackage{url}

\usepackage{epsfig}
\usepackage{graphicx}
\usepackage{amsmath}
\usepackage{amssymb}

\usepackage{amsthm}

\usepackage{algorithm2e}

\usepackage{subfigure}
\usepackage{comment}

\usepackage{physics}
\usepackage{subfigure}

\usepackage{xspace}

\usepackage{tikz}

\usepackage{float}

\usepackage{cite}

\newcommand{\codeurl}{https://anonymous.4open.science/r/sampling_matters_reproducibility-BB60/}

\begin{document}

\title{Sampling Matters in Explanations: Towards Trustworthy Attribution Analysis Building Block in Visual Models through Maximizing Explanation Certainty}

\def\preprint{1}

\if\preprint 1

    \author[*,$\dag$]{R\'{o}is\'{i}n Luo}
    \author[*,$\dag$]{James McDermott}
    \author[*,$\dag$]{Colm O'Riordan}

    \affil[*]{University of Galway (Ireland)}
    \affil[$\dag$]{SFI Centre for Research Training in Artificial Intelligence (Ireland)}

    \date{\vspace{-0.9em} j.luo2@universityofgalway.ie}
	%\date{}
\else

    \author{Anonymous Submission}
    \affil{Anonymous Affiliation}
    \date{}

\fi

\maketitle
\thispagestyle{empty}

\begin{abstract}
Image attribution analysis seeks to highlight the feature representations learned by visual models such that the highlighted feature maps can reflect the pixel-wise importance of inputs. Gradient integration is a building block in the attribution analysis by integrating the gradients from multiple derived samples to highlight the semantic features relevant to inferences. Such a building block often combines with other information from visual models such as activation or attention maps to form ultimate explanations. Yet, our theoretical analysis demonstrates that the extent to the alignment of the sample distribution in gradient integration with respect to natural image distribution gives a lower bound of explanation certainty. Prior works add noise into images as samples and the noise distributions can lead to low explanation certainty. Counter-intuitively, our experiment shows that extra information can saturate neural networks. To this end, building trustworthy attribution analysis needs to settle the sample distribution misalignment problem. Instead of adding extra information into input images, we present a semi-optimal sampling approach by suppressing features from inputs. The sample distribution by suppressing features is approximately identical to the distribution of natural images. Our extensive quantitative evaluation on large scale dataset ImageNet affirms that our approach is effective and able to yield more satisfactory explanations against state-of-the-art baselines throughout all experimental models. 

\end{abstract}

\begin{figure*}[t]

  \centering

  \resizebox{0.9\textwidth}{!}{
   \includegraphics{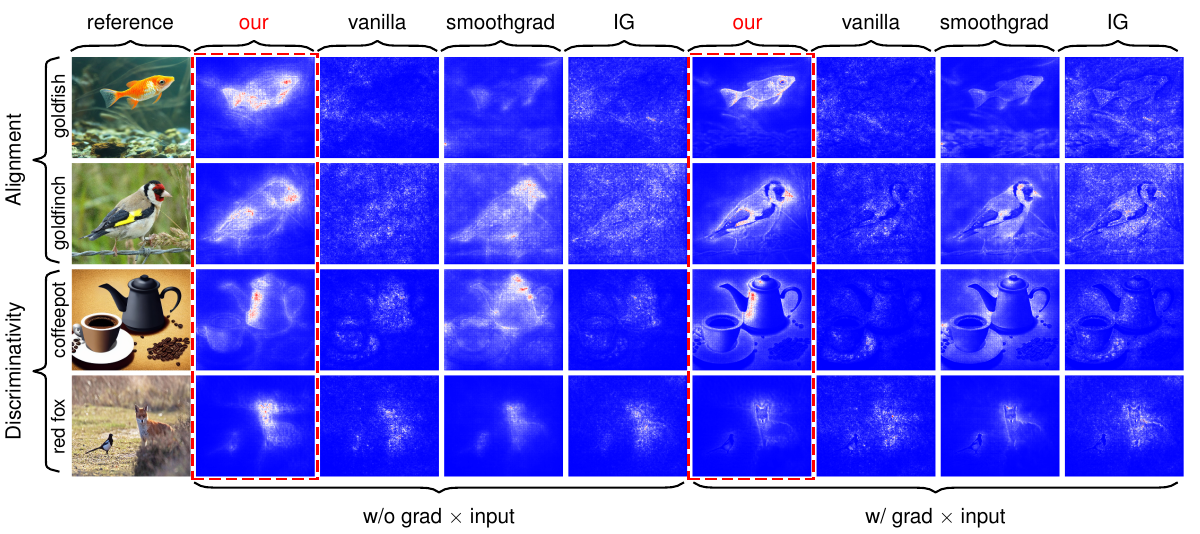} 
  }
  
  \caption{The figure is a performance showcase. The performances are qualitatively compared from two aspects: (1) The semantic alignment and (2) the object discriminativity with the presence of multiple objects. The evaluation model is a ResNet50 pre-trained on ImageNet. The iteration numbers are set to $50$. The pixel dropping probability of our algorithm is set to $0.7$. The noise level of Smoothgrad is set to $0.15$ as suggested in the original paper. The explanations are normalized to $[0,1]$ by using min-max normalization and visualized with the quantitative color map ``\href{https://matplotlib.org/stable/tutorials/colors/colormaps.html}{bwr}'' ($[\textrm{blue}=0, \textrm{white}=0.5, \textrm{red}=1]$). }

\label{fig:evaluation}
\end{figure*}

\let\svthefootnote\thefootnote
\newcommand\freefootnote[1]{%
  \let\thefootnote\relax%
  \footnotetext{\\#1}%
  \let\thefootnote\svthefootnote%
}

\freefootnote{Paper reproducibility: \url{\codeurl}.}

\section{Introduction}
\label{sec:intro}
Image semantic features are conveyed by the joint distribution of image pixels. Visual models learn how to extract the features in training process. Pixel-wise attribution analysis seeks to highlight the learned feature representations relevant to inferences as explanations \citep{linardatos2020explainable,adadi2018peeking}. Gradient integration is a major building block in attribution analysis by integrating the gradients from multiple samples sampled from some distributions \citep{simonyan2013deep,baehrens2010explain,smilkov2017smoothgrad,omeiza2019smooth,sundararajan2017axiomatic,jalwana2021cameras}. %

The first-order gradients merely provide limited local information regarding the inferences of models at a given point. Moreover, the extra information from samples can suppress the activation of networks and causes saturation (See Figure~\ref{subfig:optimal_drop_prob}). Furthermore, our experiment suggests that some gradient integration based algorithms are susceptible to the training with data augmentation. For example, fine-tuning models with data augmentation can fail the explanations (See Figure~\ref{fig:failures}). This is because the models training with data augmentation learn to ignore the samples from the sampling distributions not aligned with natural image distribution. In recent works, e.g. Grad-CAM \citep{selvaraju2017grad} and CAMERAS \citep{jalwana2021cameras}, gradient integration often serves as a building block by combining with activation or attention maps to provide more robust explanations.

Yet, our theoretical analysis suggests that such a building block in attribution analysis often suffers from low explanation certainty due to the misalignment of the sample distribution with respect to natural image distribution. Building trustworthy visual models is an imperative call for crucial scenarios such as medical imaging and autonomous driving. To this end, we must settle the sampling misalignment problem in gradient integration and seek more trustworthy sampling approach towards trustworthy visual models. 

We theoretically revisit the sampling problem from the perspective of explanation certainty and propose a semi-optimal sampling approach by suppressing features from input images. The intuition behind is that natural images densely encode information with considerable redundancy and a small fraction of pixels can still convey considerable feature information to make reasonable inferences. For example, humans can still recognize the objects in images by dropping $80\%$ of pixels (See Figure~\ref{fig:samples}).

\section{Related work}

We focus on the discussion of the sampling problem in the building block gradient integration. Therefore, we conduct a brief literature review to provide the context for this research in visual models from two aspects: (1) Local explainability and (2) gradient integration.

\begin{figure}[t]

  \centering

  \resizebox{0.68\columnwidth}{!}{
    \includegraphics[width=1\columnwidth]{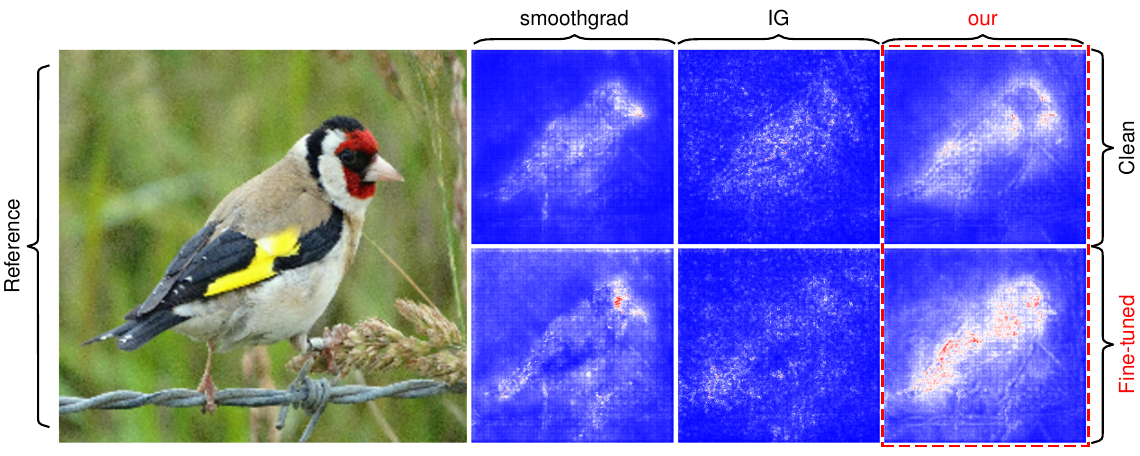} 
  }
  
  \caption{This experiment exhibits the failures and the sensitivity of the explanations by fine-tuning a pre-trained ResNet50 on ImageNet with data augmentation. The model is fine-tuned with two data augmentations: (1) Randomly adding Gaussian with $\sigma$ in $[0.1,0.3]$ and (2) randomly adjusting the luminance within $[0.1, 0.9]$. We train $5000$ batches with batch-size by $8$, learning rate $10^{-3}$ and SGD optimizer. The quality of explanations degrades due to the model learns to ignore the irrelevant perturbations in the samples by adding noise.}

\label{fig:failures}
\end{figure} 

\textbf{Local explainability}: Local explainability in visual models provides the explanations for individual images in inferences. Recent methods fall into the follow categories: 
\begin{itemize}
    %\vspace{-0.5em}
    %\setlength\itemsep{-0.2em}

    \item Local approximation based methods: e.g. LIME (Local Interpretable Model-Agnostic Explanations) \citep{mishra2017local};

    \item Shapley value theory based methods: e.g. SHAP (SHapley Additive exPlanations) \citep{lundberg2017unified};

    \item Gradient integration based methods: e.g. Smoothgrad \citep{smilkov2017smoothgrad} and IG \citep{sundararajan2017axiomatic};

    \item Activation decomposition based methods: e.g. DeepLIFT (Deep Learning Important FeaTures) \citep{shrikumar2017learning}, LRP (Layer-wise Relevance Propagation) \citep{montavon2019layer}, and CAM (Class Activation Map) \citep{li2018tell};
    
    \item Combination of gradient and activation maps: e.g. Grad-CAM \citep{selvaraju2017grad} and CAMERAS \citep{jalwana2021cameras}.
\end{itemize}

\textbf{Gradient integration}: Gradient maps capture the local semantic features from images relevant to inferences. Integrating multiple gradients from multiple samples can improve explanation quality. \citeauthor{smilkov2017smoothgrad} add noise from some normal distributions into images to form samples \citep{smilkov2017smoothgrad}. The gradients from the derived samples are then integrated to form explanations. Yet, the samples by adding random noise do not align with the distribution of natural images. Such a misalignment can yield the explanations with low explanation certainty (See the fourth and the eighth columns in Figure~\ref{fig:evaluation}). Our experiment shows that their approach can fail if visual models are trained with data augmentation as models learn to ignore the added noise (See Figure~\ref{fig:failures}). Inspired by the Aumann–Shapley theory \citep{shapley1953stochastic,roth1988shapley,aumann2015values}, \citeauthor{sundararajan2017axiomatic} attempt to tackle the neuron saturating problem by globally scaling inputs to create multiple samples from inputs. The derived multiple gradient maps from the samples are then integrated to create ultimate explanations \citep{sundararajan2017axiomatic}. Such an approach can preserve the features aligned with natural images. However, their samples suffer from the lack of feature diversity. Thus, the explanations using IG remain unsatisfactory (See the fifth and the ninth columns in Figure~\ref{fig:evaluation}). To overcome the limit that gradient maps can merely capture local semantic information, Grad-CAM \citep{selvaraju2017grad} and CAMERAS \citep{jalwana2021cameras} combine gradient maps with activation maps to form the explanations with higher quality.

\section{Sampling problem}
\label{sec:sampling_problem}

We formulate the explanation certainty and theoretically analyze the distribution alignment problem in gradient integration.

\subsection{Explanation certainty}

The explanation task for visual models asks a question for given model $\mathcal{M}$, given input image $\vb*{x}=(x_1, x_2, \cdots, x_d) \in \mathbb{R}^d$ (where $x_i$ denotes the $i$-th pixel and $d$ denotes the input dimension), given ground-truth category $y \in \mathbb{R}$ and given explanation $\vb*{z} \in \mathbb{R}^d$: To what extent, by showing the explanation $\vb*{z}$, we can infer that the explanation is more relevant to the input $\vb*{x}$. If the explanation $\vb*{z}$ and the input $\vb*{x}$ are not relevant for given model $\mathcal{M}$, we can infer that the model $\mathcal{M}$ does not align with human-specific intuitions and lack of explainability. An explanation algorithm is map: $g: (\vb*{x}, \mathcal{M}, y) \mapsto \vb*{z}$ which is expected to faithfully reflect the relevance between inputs and explanations. We define `explanation certainty' as the conditional probability $Pr(\vb*{x} | \vb*{z}; \mathcal{M}, y)$: The probability of the input $\vb*{x}$ for given explanation $\vb*{z}$, given model $\mathcal{M}$ and given ground-truth category $y$. We simplify and denote the explanation certainty as $Pr(\vb*{x} | \vb*{z})$ to keep depiction succinct thereafter.

\subsection{A lower bound of explanation certainty}
\label{sec:lower_bound}

\begin{figure}[!t]

  \centering

  \resizebox{0.96\columnwidth}{!}{
  \includegraphics[width=1\columnwidth,height=0.143\columnwidth]{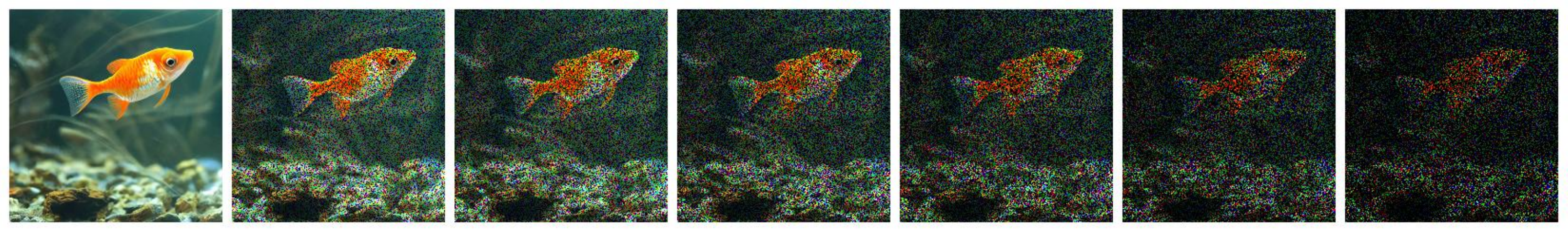} 
  }
  \caption{This example shows the information redundancy in nature images. The leftmost is the reference. The images from the rightmost to leftmost aside the reference are with random pixel sampling ratio by $0.7$, $0.6$, $0.5$, $0.4$, $0.3$ and $0.2$ respectively.
  }

\label{fig:samples}
\end{figure}

It is not difficult to show that the mutual information \citep{kullback1997information,cover1999elements} between inputs and its explanations is a lower bound of $Pr(\vb*{x} | \vb*{z})$. Let further assume natural images are pixel-wise i.i.d. to simplify the theoretical analysis. Let $p(x|z)$ be the conditional p.d.f. for some pixel $x \in \vb*{x}$ at given some pixel $z \in \vb*{z}$. Hence the $Pr(\vb*{x} | \vb*{z})$ can be approximately rewritten as:
\begin{align}
    \label{equ:def_px_cond_z}
    Pr(\vb*{x} | \vb*{z}) &= 
    \prod\limits_x \prod\limits_z Pr(x|z)
    = \exp(\sum\limits_x \sum\limits_z \log Pr(x|z)) %
    \approx  \exp(\iint_{z,x} \log p(x|z) dxdz).
\end{align}
Considering:
\begin{align}
    \label{equ:prob_identity}
    \iint_{z,x}  p(x,z) dxdz = 1
\end{align}
and the mutual information between $\vb*{x}$ and $\vb*{z}$:
\begin{align}
    \label{equ:def_ixz}
    I(\vb*{x}; \vb*{z}) = \iint_{z,x} p(x, z) \log (\frac{p(x|z)}{p(x)}) dxdz
\end{align}
and the entropy of $\vb*{x}$:
\begin{align}
    \label{equ:def_hx}
    H(\vb*{x}) = -\int_x p(x)\log p(x) dx.
\end{align}

Combining above results in equations (\ref{equ:def_px_cond_z}, \ref{equ:prob_identity}, \ref{equ:def_ixz} and \ref{equ:def_hx}) and applying H\"{o}lder's inequality:
\begin{align}
    \iint_{z,x} \log p(x|z) dxdz &= 1 \cdot \iint_{z,x} \log p(x|z) dxdz = \iint_{z,x} p(x, z) dxdz \cdot \iint_{z,x} \log p(x|z) dxdz \nonumber \\
    &\geq \iint_{z,x} p(x, z) \log (\frac{p(x|z)}{p(x)}p(x)) dxdz \nonumber \\
    &= \iint_{z,x} p(x, z) \log (\frac{p(x|z)}{p(x)}) dxdz + \int_{x} (\int_{z} p(x,z)dz) \log p(x) dx \nonumber\\
    &= I(\vb*{x}; \vb*{z}) + \int_{x} p(x) \log p(x) dx \nonumber\\
    &= I(\vb*{x}; \vb*{z}) - H(\vb*{x}) \equiv -H(\vb*{x}|\vb*{z}).
\end{align}
Hence mutual information gives a lower bound of explanation certainty by:
\begin{align}
    \label{equ:mi_and_mle}
    Pr(\vb*{x} | \vb*{z}) \geq \frac{\exp(I(\vb*{z}; \vb*{x}))}{\exp(H({\vb*{x}}))} \equiv \exp(-H(\vb*{x}|\vb*{z})). ~\qed
\end{align}

\subsection{Gradient integration}

\begin{figure}[t!] %
	\centering

	\subfigure[Sample distribution alignment experiment.]{
		\begin{minipage}[b]{0.43\textwidth}
			\includegraphics[width=1\textwidth,height=1\textwidth]{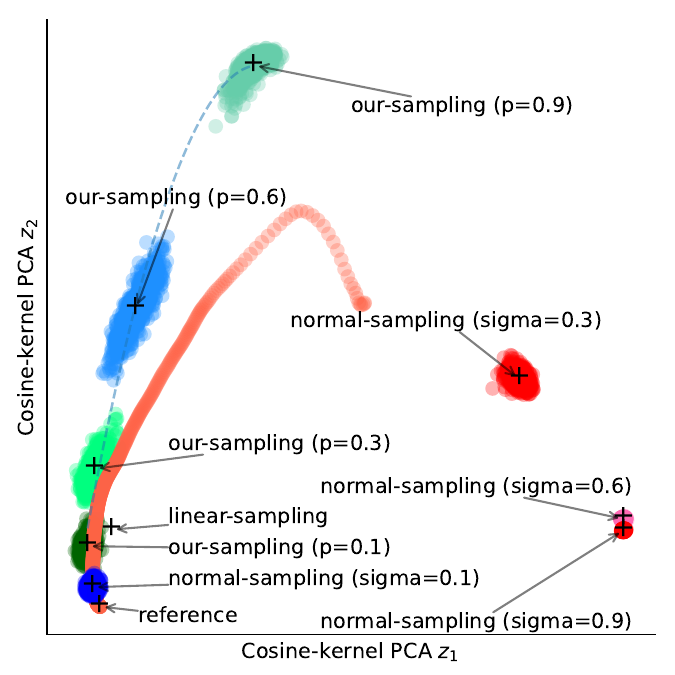}
		\end{minipage}
		\label{subfig:feature_projections}
	}\hspace*{\fill}%
	\subfigure[Network activation experiment and mutual information experiment with respect to various various pixel dropping probability $p$.]{
		\begin{minipage}[b]{0.43\textwidth}

        \centering

        \includegraphics[width=1\textwidth,height=1\textwidth]{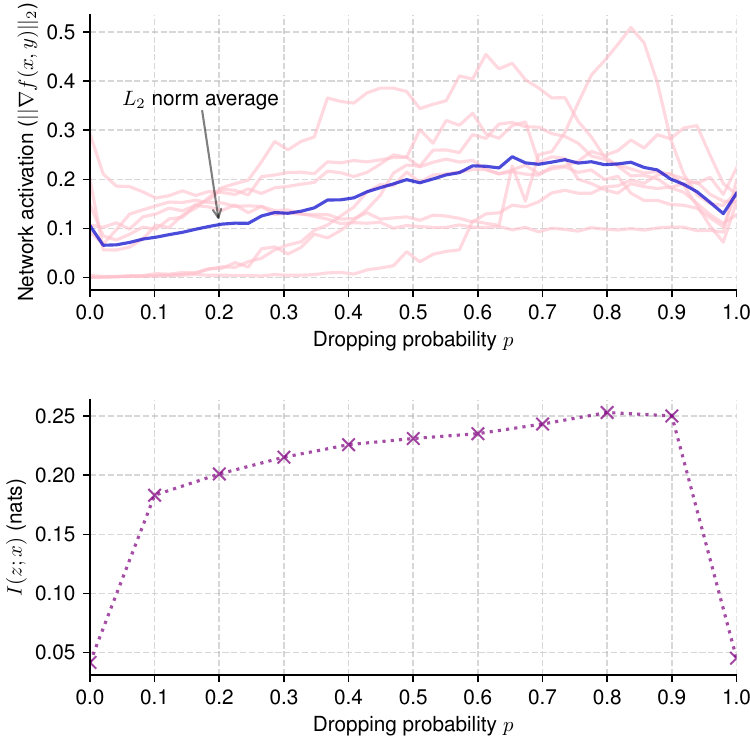} 
        
		\end{minipage}
		\label{subfig:optimal_drop_prob}
	}

	\caption{In the left figure we investigate how neural networks respond to various sampling distributions. We use a ResNet50 pre-trained on ImageNet and project the embeddings from the penultimate layer with cosine-kernel PCA. We collect $1000$ samples for each sampling approach. We vary the noise level of normal-sampling (used in Smoothgrad) from $0.1$ to $0.9$. We vary the pixel dropping probability of our-sampling approach from $0.1$ to $0.9$. We use ``$+$'' to indicate the projection centers. In the right figure, we empirically investigate how neural networks activate by measuring the average gradient norms with respect to various pixel dropping probability $p$ in our approach. We also conduct the experiment to measure how the dropping probability $p$ affects explanation certainty (mutual information). The result shows that the optimal dropping probability $p$ is from $0.6$ to $0.8$.}
	
	\label{fig:feature_projections_and_optimal_prob}
	
\end{figure}

The gradient integration without the technique of `$\mathrm{inputs} \times \mathrm{gradients}$' \citep{shrikumar2017learning,sundararajan2017axiomatic} can be formulated as:
\begin{align}
\label{equ:gradient_integration}
    \vb*{z}  := %
    \mathop{\mathbb{E}}_{\hat{\vb*{x}} \sim q(\vb*{x})} \left[|\nabla_{\hat{\vb*{x}}} f(\hat{\vb*{x}},y)| \right]
\end{align}
where $|\cdot|$ denote point-wise absolute value modulus, $\vb*{x}$ denotes some input, $\vb*{z}$ denotes some explanation, $\hat{\vb*{x}}$ denotes some sample from some sampling distribution $q(\vb*{x})$, $y$ denotes ground-truth label, and $f: (\vb*{x}, y) \mapsto \mathbb{R}$ defines some image classifier. 

\subsection{Optimal sampling}

Explanation certainty is lower bounded by the mutual information between inputs and explanations for given models. Maximizing the explanation certainty in gradient integration based approach can improve the trustworthiness and faithfulness of explanations. Let $q_{\vb*{x}}$ be some sampling distribution for given input $\vb*{x}$ in gradient integration. Let $p$ be natural image distribution. Let $q^*$ be the optimal sampling distribution. We have:
\begin{align}
    \label{equ:optimal_sampling}
    q^* 
    = \arg\max\limits_q \mathop{\mathbb{E}}_{\vb*{x} \sim p} \left[ I(\vb*{x}, \vb*{z}) \right] %
     = \arg\max\limits_q \mathop{\mathbb{E}}_{\vb*{x} \sim p} \left[ I(\vb*{x}, \mathop{\mathbb{E}}_{\hat{\vb*{x}} \sim q_{\vb*{x}}} [|\nabla_{\hat{\vb*{x}}} f(\hat{\vb*{x}},y)|]) \right] %
\end{align}

Considering that mutual information function $I(\vb*{x}, \vb*{z})$ is convex for given $\vb*{x}$ and applying Jensen's inequality:
\begin{align}
    \label{equ:sampling_matters}
    I(\vb*{x}, \vb*{z}) 
    = I(\vb*{x}; \mathop{\mathbb{E}}_{\hat{\vb*{x}} \sim q_{\vb*{x}}} [|\nabla_{\hat{\vb*{x}}} f(\hat{\vb*{x}},y)|]) %
    \leq \mathop{\mathbb{E}}_{\hat{\vb*{x}} \sim q_{\vb*{x}}} [I(\vb*{x}; |\nabla_{\hat{\vb*{x}}} f(\hat{\vb*{x}},y)|)] %
    = \mathop{\mathbb{E}}_{\hat{\vb*{x}} \sim q_{\vb*{x}}} [I(\vb*{x};\hat{\vb*{z}})].
\end{align}

By using equation (\ref{equ:sampling_matters}), we now rewrite the equation (\ref{equ:optimal_sampling}) into:
\begin{align}
    \label{equ:optimal_sampling_rewrite}
    q^* = \arg\max\limits_q \mathop{\mathbb{E}}_{\vb*{x} \sim p_{\vb*{x}}} \left[ \mathop{\mathbb{E}}_{\hat{\vb*{x}} \sim q_{\vb*{x}}} [I(\vb*{x};\hat{\vb*{z}}(\hat{\vb*{x}}))] \right]
\end{align}
which has the optimal solution $q^*=p$. This result implies that if samples come from the same distribution as natural images. The mutual information in gradient integration between inputs and explanations will be maximized.

\section{Sampling with feature suppression}

Natural images encode information densely with considerable redundancy (Figure~\ref{fig:samples}). Humans can still give reasonable inferences for images when a majority of pixels is dropped. Unlike that prior works add extra noise into images to create samples in gradient integration, we instead remove information from images. Interestingly, in our experiment, we have also observed that less information by dropping a proper proportion of pixels can further increase the activation of networks and maximize the explanation certainty. Figure~\ref{subfig:optimal_drop_prob} provides the experiments to show such an observation. 

\subsection{Feature suppression}
For each image $\vb*{x} = (x_i)_{i=1}^{d}$ with $d$ pixels, we sample an image $\vb*{x}^{*}$ from $\vb*{x}$ by doing pixel-wise Bernoulli trial for the $i$-th pixel from set $\{x_i, 0\}$ with dropping probability $p$. Thus $x_i^{*} = \mathbb{B}(\{x_i, 0\}; 1 - p)$ where $\mathbb{B}$ gives a Bernoulli trial with probability $1-p$. Our approach is formulated as:
\begin{align}
\label{equ:our_approach}
    \vb*{z}  := \mathop{\mathbb{E}}_{\hat{\vb*{x}} \sim \mathbb{B}(\{\vb*{x}, 0\}; 1 - p)} \left[|\nabla_{\hat{\vb*{x}}} f(\hat{\vb*{x}},y)| \right]
\end{align}
where $\mathbb{B}(\{\vb*{x}, 0\}; 1 - p)$ denotes our sampling operation, $\hat{\vb*{x}}$ denotes a sample, $\vb*{z}$ denotes the explanation, $y$ denotes some ground-truth label, and $f: (\hat{\vb*{x}}, y)$ denotes some model.

\subsection{Sampling alignment experiment}

The experiment in Figure~\ref{subfig:feature_projections} shows how neural networks respond to samples by taking the embeddings from the penultimate layer of a ResNet50 pre-trained on ImageNet and projecting the embeddings with cosine-kernel PCA. The `linear-sampling' refers to simply scale the pixel intensities of images in global. The distribution of the `linear-sampling' aligns with the distribution of natural images. The `normal-sampling' refers to the samples derived by adding the noise from various normal distributions. The result shows that our samples (marked as `our-sampling') align with the natural images. The samples from `normal-sampling' do not align well with natural images. The projections show that models treat the samples from various distributions with different patterns. %

\subsubsection{Optimal $p$}

We empirically investigate the optimal dropping probability $p$ in Figure~\ref{subfig:optimal_drop_prob} by: (1) Measuring how networks activate with respect to various $p$ and (2) measuring how explanation certainty (mutual information) changes with respect to various $p$. We vary $p$ from $0$ to $1$ with stride $0.1$. All results are measured on a ResNet50 pre-trained on ImageNet. The result shows that when $p$ falls in $[0.6, 0.8]$ the network activation will be maximized and the mutual information will be maximized as well. We also repeat this experiment on other architectures. The pattern remains the same. There are two findings which are counter-intuitive: (1) When we suppress information from inputs, the activations of networks first increase and networks extract more relevant features, and, (2) as we continue to suppress more information, networks become more confusing and less activated and fail to extract relevant features. This experiment can further justify our approach by suppressing features from inputs. The behind mechanism still needs further investigation.

\begin{figure*}[!t]

  \centering

  \resizebox{0.8\textwidth}{!}{
  \includegraphics[width=1\textwidth,]{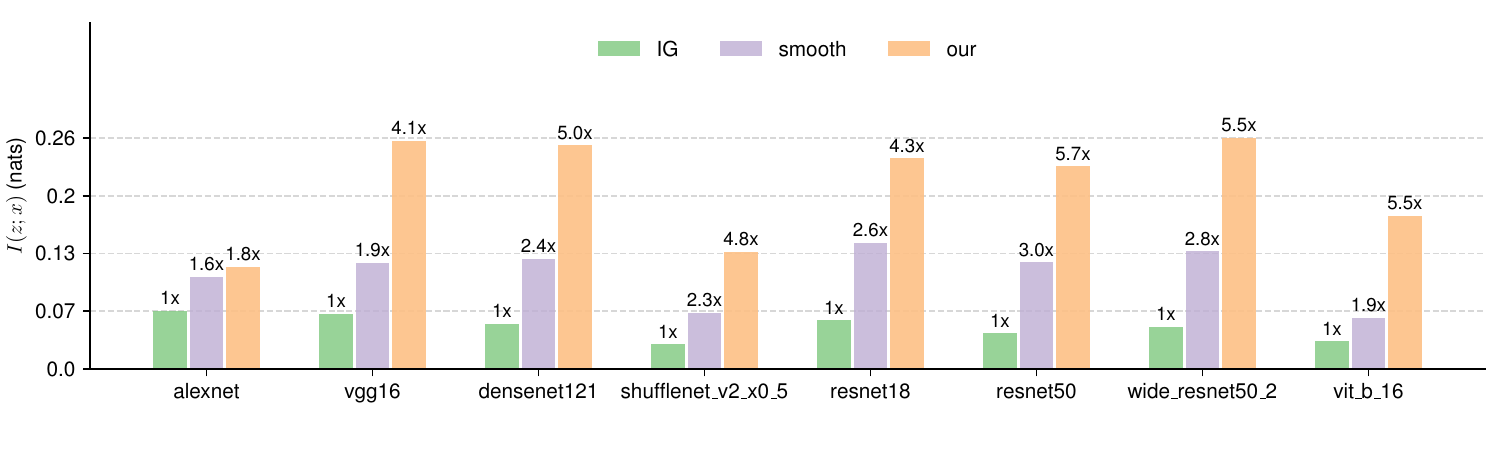} 
  }
  
  \caption{We extensively evaluate our algorithm against the gradient integration baselines over ImageNet dataset. All random seeds are set to $1860867$ to guarantee the reproducibility. We choose the models pre-trained on ImageNet. We use $100$ epochs and $100$ images for all methods. For all chosen models, our approach outperforms state-of-the-art baselines.}

\label{fig:scores}
\end{figure*}

\section{Evaluation}

As we focus on investigating the sampling problem in the building block for attribution analysis, we set three baselines: Smoothgrad, IG and vanilla due the research scope. We perform both qualitative and quantitative evaluations. %

\subsection{Qualitative showcase} 
In Figure~\ref{fig:evaluation}, we showcase the results with and without the `$\mathrm{Gradient} \times \mathrm{Input}$' technique. We choose a ResNet50 pre-trained on ImageNet. The experiment examines the performance from two aspects: (1) Semantic feature alignments and (2) the object discriminativity with the presence of multiple objects. The result shows that our approach can yield more satisfactory explanations intuitively. 

\subsection{Quantitative evaluation}

We also conduct an extensive quantitative evaluation by measuring the lower bound (mutual information) of explanation certainty. We choose  multiple models pre-trained on ImageNet. The mutual information unit is in \textit{nats} and the results are also normalized with respect to the IG on the model basis. The images are randomly sampled from ImageNet by fixing random seed for reproducibility purpose. The result shows that our approach outperform all baselines for all models.

\section{Conclusions}
We theoretically show that mutual information between explanations and inputs gives a lower bound of explanation certainty in the the explanation building block for image models with gradient integration approach. Maximizing explanation certainty can achieve trustworthy and faithful explanations. Due to the discussion scope, we have not unfolded: (1) The analysis in terms of the sensitivity to the training with data augmentation and (2) the results when our approach is incorporated with other information such as class activation or attention maps. A further investigation is necessary to address such concerns.

\if\preprint 1
\section*{Acknowledgments}
This publication has emanated from research [conducted with the financial support of/supported in part by a grant from] Science Foundation Ireland under Grant number 18/CRT/6223. For the purpose of Open Access, the author has applied a CC BY public copyright licence to any Author Accepted Manuscript version arising from this submission. We also thank reviewers for their constructive comments which can significantly improve our research quality. We thank the support from the ICHEC (Irish Centre for High-End Computing). We also thank the contributor Jiarong Li for the help in the discussions and proofreading. 
\fi

\bibliographystyle{apalike}

\bibliography{ref}

\end{document}